\documentclass[10pt,twoside]{article}

\usepackage[T1]{fontenc}
\usepackage[left,mathlines]{lineno}
\usepackage[dvipsnames]{xcolor}
\usepackage[width=122mm,left=12mm,paperwidth=146mm,height=193mm,top=12mm,paperheight=217mm]{geometry}
\usepackage{amsmath}
\usepackage{amssymb}
\usepackage{algpseudocode}
\usepackage{cite}
\usepackage{xspace}
\usepackage[hyphens]{url}
\usepackage{etoolbox}
\usepackage[labelfont=bf,font=small,tableposition=bottom]{caption}
\usepackage[skip=3pt]{subcaption}
\definecolor{accvblue}{rgb}{0,0,1.0}
\usepackage{graphicx}
\usepackage{booktabs}
\usepackage{array}
\usepackage{makecell}
\usepackage{standalone}
\usepackage[accsupp]{axessibility}
\usepackage[pagebackref,breaklinks,colorlinks,citecolor=accvblue,hypertexnames=false]{hyperref}
\usepackage[capitalize]{cleveref}

\makeatletter
\newcount\cv@tmpc@
\newcount\cv@tmpc
\def\fillzeros[#1]#2{\cv@tmpc@=#2\relax\ifnum\cv@tmpc@<0\cv@tmpc@=-\cv@tmpc@\fi
\cv@tmpc=1 \loop\ifnum\cv@tmpc@<10 \else \divide\cv@tmpc@ by 10 \advance\cv@tmpc by 1 \fi
   \ifnum\cv@tmpc@=10\relax\cv@tmpc@=11\relax\fi \ifnum\cv@tmpc@>10 \repeat
\ifnum#2<0\advance\cv@tmpc1\relax-\fi
\loop\ifnum\cv@tmpc<#1\relax0\advance\cv@tmpc1\relax\fi \ifnum\cv@tmpc<#1 \repeat
\cv@tmpc@=#2\relax\ifnum\cv@tmpc@<0\cv@tmpc@=-\cv@tmpc@\fi \relax\the\cv@tmpc@}

\newcommand*\linenomathpatch[1]{\cspreto{#1}{\linenomath}\cspreto{#1*}{\linenomath}\csappto{end#1}{\endlinenomath}\csappto{end#1*}{\endlinenomath}}
\newcommand*\linenomathpatchAMS[1]{\cspreto{#1}{\linenomathAMS}\cspreto{#1*}{\linenomathAMS}\csappto{end#1}{\endlinenomath}\csappto{end#1*}{\endlinenomath}}

\expandafter\ifx\linenomath\linenomathWithnumbers
  \let\linenomathAMS\linenomathWithnumbers
  \patchcmd\linenomathAMS{\advance\postdisplaypenalty\linenopenalty}{}{}{}
\else
  \let\linenomathAMS\linenomathNonumbers
\fi

\linenomathpatch{equation}
\linenomathpatchAMS{gather}
\linenomathpatchAMS{multline}
\linenomathpatchAMS{align}
\linenomathpatchAMS{alignat}
\linenomathpatchAMS{flalign}

\patchcmd{\mmeasure@}{\measuring@true}{
  \measuring@true
  \ifnum-\linenopenaltypar>\interdisplaylinepenalty
    \advance\interdisplaylinepenalty-\linenopenalty
  \fi
  }{}{}
\makeatother

\makeatletter
\DeclareRobustCommand\onedot{\futurelet\@let@token\@onedot}
\def\@onedot{\ifx\@let@token.\else.\null\fi\xspace}
\makeatother

\newcommand{\keywords}[1]{\par\smallskip\noindent\textbf{Keywords: }{\def\and{\unskip; }#1}}

\date{}

\newcommand{\authorblock}[3]{\begin{minipage}[t]{#1}
    \centering
    \textbf{#2}\\[0.2em]
    #3
  \end{minipage}}

\makeatletter
\renewcommand{\maketitle}{\par
\noindent\rule{\textwidth}{2.5pt}\par
  \vspace{1.2em}
{\centering
    \Large\bfseries \@title\par
  }
  \vspace{1.2em}
\noindent\rule{\textwidth}{0.5pt}\par
  \vspace{2.2em}
  \noindent
  \makebox[\textwidth][c]{\@author}\par
  \vspace{1.5em}
}
\makeatother

\newcolumntype{C}{>{\centering\arraybackslash}p{1.75cm}}

\newcommand{\ccsconcepts}[1]{\par\smallskip\noindent\textbf{CCS CONCEPTS} #1\par}

\title{A Study of Hidden-State Optimization Order in Predictive Coding Networks}
\author{
  \authorblock{0.48\textwidth}
    {Xueyuan Li}{Kyushu Universit}\hfill
  \authorblock{0.48\textwidth}
    {Danilo V. Vargas}{Kyushu University}
}

\begin{document}

\maketitle

\begin{abstract}
Local learning methods offer an alternative to end-to-end backpropagation, but their unstructured local objectives can produce weak feature learning in deep networks. We study whether the order of hidden-state optimization can address this limitation. We propose a boundary-first inference schedule that partitions a model into chunks, first coordinates hidden states at chunk boundaries, and then refines representations within each chunk. We instantiate this schedule in predictive coding networks (PCNs), a local-learning framework in which hidden activities and prediction errors are explicitly exposed during inference. On CIFAR-10, PC-PC improves accuracy over standard PC by 9.77 percentage points under a standard parametrization, while the PC-BP hybrid improves accuracy over standard PC by 5.51 percentage points under a $\mu$-parametrization. Diagnostic analyses further show more non-trivial early-layer updates, lower initial-to-final CKA, and more diverse layerwise gradients, consistent with stronger feature learning. These results support boundary-first, chunk-based inference as a practical design principle for predictive-coding training and motivate its study in broader local-learning systems.
\end{abstract}
\ccsconcepts{
\textbullet\ Computing methodologies $\rightarrow$ Neural networks;
Learning latent representations;
Bio-inspired approaches}


\section{Introduction}
Deep networks learn useful representations when error signals produce informative updates throughout the model. End-to-end backpropagation achieves this by propagating a global supervised objective through all layers, while many alternative learning systems try to replace this global dependency with local objectives and local state updates. Their properties, including biological plausibility, modular computation, cortical-like inference, and local error minimization, are attracting the attention of both neuroscientists and machine learning researchers \cite{Rao1999, Whittington2017, Scellier2017, Bengio2014, Belilovsky2019, Xiong2020}.

The same locality can also become a bottleneck. When learning signals are transmitted only through adjacent objectives, early and intermediate layers may receive weak or degenerate updates. Predictive Coding Networks (PCNs) provide a clear example of this failure mode (\cref{fig:grad-similarity-minist}). 
Although PCNs can be trained by minimizing local prediction errors, recent work has reported imbalanced error propagation, weak early-layer updates, and unstable inference dynamics \cite{Ha2026, Pinchetti2025, Alonso2024, Salvatori2024a, Frieder2022}. Stabilization methods such as residual connections, layerwise normalization, and Maximal Update Parametrization improve optimization in deep PCNs \cite{Innocenti2025, Yang2020}; however, how local objectives should be organized to encourage feature learning remains an open question.

\begin{figure}[t!]
    \centering
    \includegraphics[width=0.8\textwidth]{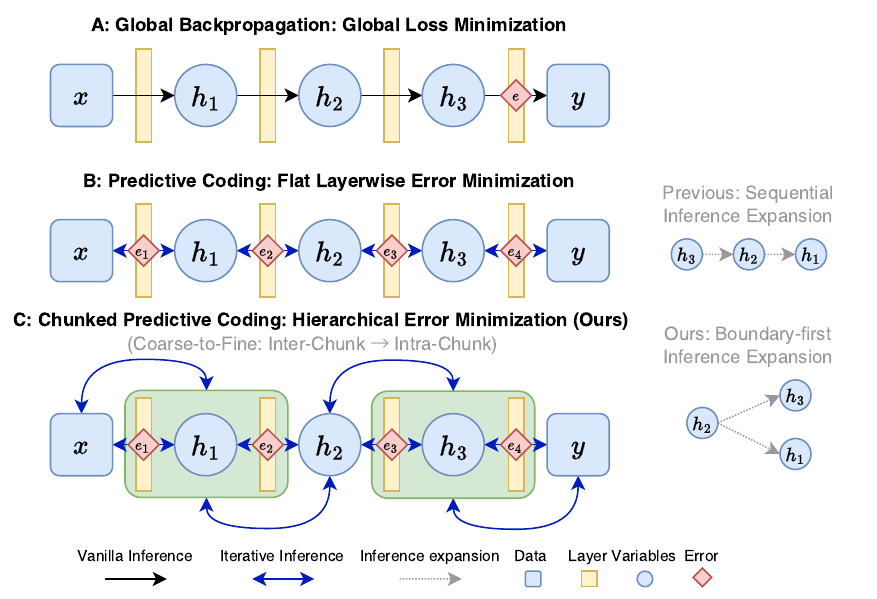}
    \caption{Illustration of global backpropagation, standard predictive coding, and boundary-first predictive coding with feedforward initialization. (A) Global backpropagation computes a supervised loss at the output and propagates gradients through the full network. (B) Standard predictive coding minimizes local prediction errors between adjacent layers, with the inferred activity set expanding from $\{h_3\}$ to $\{h_2,h_3\}$ and then to $\{h_1,h_2,h_3\}$. (C) Boundary-first predictive coding uses inter-chunk inference to coordinate boundary states before intra-chunk inference refines activities within each chunk; the inferred activity set expands from $\{h_2\}$ to $\{h_1,h_2,h_3\}$.}
    \label{fig:overall}
\end{figure}

Our work studies a different route: changing the order in which hidden activities are optimized rather than only changing the objective parametrization. We propose a boundary-first inference schedule that first coordinates hidden states at chunk boundaries and then refines the remaining states within each chunk. Chunking defines the boundary and interior variable sets that make this optimization order operational (\cref{fig:overall}).

We instantiate this schedule in predictive coding because PCNs already maintain hidden activities and prediction errors during inference. The instantiation first performs inter-chunk inference to align boundary activities, then performs intra-chunk inference to refine local representations. We refer to the fully predictive-coding version as PC-PC. We also evaluate PC-BP as a hybrid control, where chunk boundaries are coordinated by predictive-coding inference but intra-chunk learning uses backpropagation. This control tests whether the gains of the boundary-first, chunk-based schedule persist when the within-chunk learning rule changes.

Our contributions center on inference expansion order in local learning systems:
\begin{itemize}
    \item We formulate a boundary-first hidden-state optimization schedule for local-error training systems. We instantiate this schedule in predictive coding by using chunks to define boundary and interior states, first aligning the boundary states through inter-chunk inference and then refining the remaining activities through intra-chunk inference.

    \item We show that the boundary-first predictive-coding instantiation enables all layers (20/20) to receive non-trivial updates, whereas standard PC updates only 4/20 layers in a standard Multilayer Perceptron (MLP) and 5/20 layers in a $\mu$MLP. On CIFAR-10, PC-PC improves accuracy over standard PC by 9.77 percentage points under a standard parametrization, while the PC-BP hybrid improves accuracy over standard PC by 5.51 percentage points under a $\mu$-parametrization.

    \item We analyze representation learning with initial-to-final CKA and feature diversity with layerwise gradient similarity. The results show that PC-PC and PC-BP reduce CKA relative to standard PC and produce lower gradient similarity than global backpropagation in the measured settings.

\end{itemize}

We study this optimization schedule in predictive coding networks. Extending boundary-first inference to other local-learning frameworks is left for future work.

\section{Related Work}

\textbf{Predictive Coding Networks.} Predictive coding networks train deep models by minimizing prediction errors between adjacent layers. Early work showed that local Hebbian-style plasticity in PCNs can approximate error backpropagation \cite{Rao1999, Whittington2017}, and recent studies developed PCNs as a broader learning framework \cite{Salvatori2021, Alonso2022, Seely2025}. The main obstacle for deep PCNs is not only whether local learning is possible, but also whether the local objectives produce informative updates across depth. Maximal Update Parameterised PCNs ($\mu$PCNs) address part of this issue by stabilizing the initialization of predictive-coding inference \cite{Innocenti2025, Yang2020}. Our work is complementary to these approaches: we ask how the order of hidden-state optimization affects predictive-coding training across depth.

\textbf{Chunking and modular training.} Chunking groups low-level elements into higher-level units. In cognition, chunking explains how structured units can expand effective working-memory capacity \cite{Miller1956}. In machine learning, related ideas appear in feedforward training, including blockwise learning, modular training, and greedy layerwise optimization. Greedy layerwise learning showed that auxiliary local subproblems can train deep networks, and that multi-layer blocks can scale better than single-layer objectives \cite{Belilovsky2019}. These results motivate chunking as a way to define groups of variables. Our work uses these groups to impose a boundary-first optimization order during iterative inference in local learning systems.

\textbf{Feature learning and kernel learning.} Prior work distinguishes feature learning from kernel-like regimes in which the effective neural tangent kernel changes little during training \cite{Yang2021, Littwin2023}. The learning dynamics can be written as
\begin{equation}
    f_{t+1}(\xi) = f_t(\xi) - \eta K_t(\xi, \xi_t) \mathcal{L}'(f_t(\xi_t), y_t)
    \label{equ:kernel-feature}
\end{equation}
where $f_t$ is the model at training step $t$, $(\xi_t,y_t)$ is a training sample pair, $\eta$ is the learning rate, $\mathcal{L}'(\cdot)$ is the loss derivative, and $K_t(\xi, \xi_t)$ is the neural tangent kernel
\begin{equation}
    K_t(\xi, \xi_t) = \langle \nabla_\theta f_t(\xi), \nabla_\theta f_t(\xi_t) \rangle .
    \label{equ:ntk}
\end{equation}
If $K_t$ remains close to $K_0$, learning behaves like kernel learning with a nearly fixed representation. If $K_t$ changes during training, the model is regarded as learning features. We use Centered Kernel Alignment (CKA) to measure the similarity between initial and final representations \cite{Kornblith2019}:
\begin{equation}
    \mathrm{CKA}(H_0,H_t)=\frac{\left\|H_0^{\top}H_t\right\|_F^2}{\left\|H_0^{\top}H_0\right\|_F\left\|H_t^{\top}H_t\right\|_F}
    \label{equ:cka}
\end{equation}
where $H_0$ and $H_t$ are representations at initialization and at training step $t$. Lower CKA indicates greater representation change.

\textbf{Feature diversity.} Feature diversity also depends on whether layers receive distinct update directions. Previous work has noted that gradient diversity can support feature diversity \cite{Yang2024}. If adjacent layers have similar backward gradients and similar forward activations, their parameter updates can become redundant. We measure layerwise update diversity using the cosine similarity between weight gradients:
\begin{equation}
    \mathcal{G}_S(i,j) = \frac{\langle g_i, g_j \rangle}{\|g_i\| \|g_j\|}
    \label{equ:gradient-similarity}
\end{equation}
where $g_i$ and $g_j$ are the gradients of layers $i$ and $j$. Lower similarity indicates more diverse layerwise updates.

\section{Methodology}
\subsection{Standard Predictive Coding Training}

Predictive coding is a natural testbed for hidden-state optimization order because it represents hidden activities explicitly and updates them through local prediction errors. Consider an $N$-layer network with activities $\{h_i\}_{i=0}^{N}$, where $h_0=x$ is the input and $h_N$ is the output. For each layer $i\in\{1,\ldots,N\}$, define the forward mapping
\begin{equation}
\hat{h}_i = f_i(h_{i-1};W_i) \;\;:=\;\; \sigma_i(W_i h_{i-1} + b_i),
\end{equation}
where $W_i$ are weights, $b_i$ are biases, and $\sigma_i(\cdot)$ is a nonlinearity. The value $\hat{h}_i$ is the prediction of the $i$-th layer activity.

A discriminative PCN minimizes local prediction errors together with a supervised output loss during inference:
\begin{equation}\label{equ:prediction-error}
\text{Inference Step:}\min_{\{h_i\}_{i=1}^{N-1}}\;\sum_{i=1}^{N-1} e_i(h_i,\hat{h}_i) \;+\; \mathcal{L}(\hat{h}_N, y),
\end{equation}
where $e_i(\cdot,\cdot)$ measures the mismatch between activity $h_i$ and prediction $\hat{h}_i$. A common choice is $e_i(h_i,\hat{h}_i)=\tfrac{1}{2}\|h_i-\hat{h}_i\|_2^2$. With weights fixed, hidden activities are updated by gradient descent on \cref{equ:prediction-error}. Each activity update depends on neighboring layers through local error terms. With feedforward initialization, the output-driven error signal propagates sequentially toward earlier layers over successive inference steps, although the activity-update rule is applied to all hidden activities at every step. After inference, PCN training updates weights using the inferred activities:
\begin{equation}
\text{Learning Step:}\quad W_i \leftarrow
\begin{cases}
W_i - \eta \nabla_{W_i} e_i(h_i,\hat{h}_i),
& 1 \leq i < N, \\[0.5em]
W_i - \eta \nabla_{W_i}\mathcal{L}(\hat{h}_i,y),
& i = N.
\end{cases}
\label{equ:weight-update}
\end{equation}
Here $\eta$ is the learning rate and $\nabla_{W_i}$ denotes the gradient with respect to $W_i$. During the weight update, all inferred activities are held fixed, with $\hat{h}_i=f_i(h_{i-1};W_i)$. Before inference begins, activities are initialized with a feedforward pass, following common PCN practice \cite{Sennesh2024, Tscshantz2023, Salvatori2024, Zwol2024, Millidge2024}.

\subsection{Chunk-Based Boundary-First Predictive-Coding Inference}

\begin{figure}[t!]
    \centering
    \includegraphics[width=0.8\textwidth]{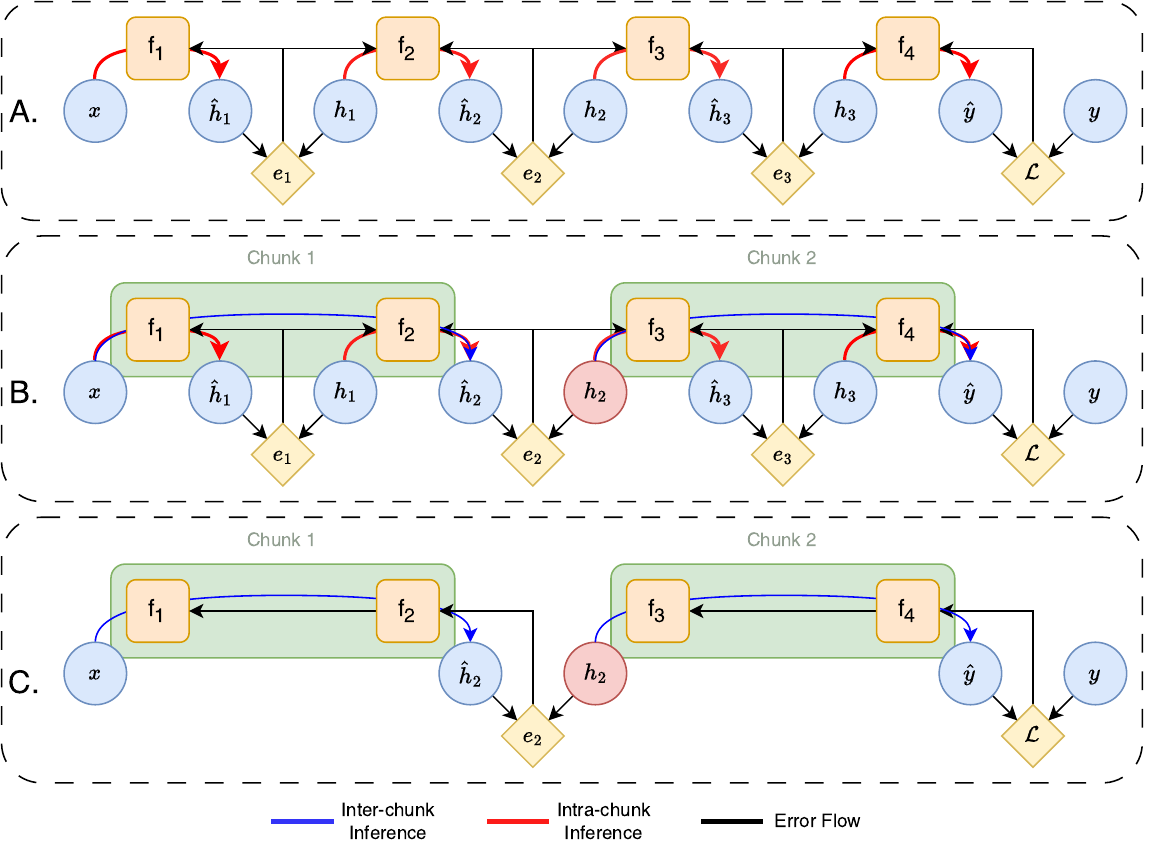}
    \caption{Inference organization in standard predictive coding and boundary-first predictive coding. Panel A shows standard predictive coding, where the update rule is applied to all hidden activities at each inference step using adjacent prediction errors. With feedforward initialization, nonzero updates propagate from the output toward earlier layers over successive steps, producing the sequential expansion illustrated in \cref{fig:overall}. Panel B shows PC-PC, where boundary states are first coordinated by inter-chunk prediction errors and hidden activities are then refined within chunks. Panel C shows PC-BP, a hybrid control in which boundary coordination is retained while learning inside each chunk uses backpropagation.}
    \label{fig:chunk-pc}
\end{figure}

In standard PC, the output-driven error signal propagates
sequentially toward earlier layers through adjacent-layer
errors. The proposed schedule reorganizes this propagation
by first coordinating chunk boundaries and then refining
internal activities. Chunking defines two sets of variables: boundary states and non-boundary states. The procedure then has two stages. First, inter-chunk inference coordinates boundary states between chunks. Second, intra-chunk inference refines the remaining hidden activities inside each chunk (\cref{fig:chunk-pc}).

Let $\mathcal{B}$ denote the set of boundary states between chunks ($\mathcal{B} = \{h_2\}$ in \cref{fig:chunk-pc}). The boundary states are updated by solving
\begin{equation}
    \text{Inter-chunk Inference:}\quad
    \min_{\{h_j \mid 1\leq j<N,\;h_j\in\mathcal{B}\}}
    \sum_{\substack{j=1\\h_j\in\mathcal{B}}}^{N-1}
    e_j(h_j,\hat{h}_j)
    + \mathcal{L}(\hat{h}_N,y)
    \label{equ:inter-chunk}
\end{equation}
where $\hat{h}_j$ is the prediction of the boundary state $h_j$ from the adjacent chunk. The set $\mathcal{B}$ contains only hidden activities at boundaries between chunks, excluding the input and output. For a chunk spanning layers $a+1,\ldots,j$, its prediction is $\hat{h}_j=(f_j\circ\cdots\circ f_{a+1})(h_a)$, where $h_a$ is the preceding boundary activity, or $h_0=x$ for the first chunk. The final chunk similarly predicts $\hat{h}_N$. During inter-chunk inference, only activities in $\mathcal{B}$ are optimized; weights, input $x$, and target $y$ remain fixed, as in standard predictive coding training.

For the fully predictive-coding instantiation, intra-chunk refinement updates non-boundary activities ${\{h_k \mid h_k \notin \mathcal{B}\}}$ ($h_1$ and $h_3$ in \cref{fig:chunk-pc}) by
\begin{equation}
    \text{Intra-chunk PC Inference:}\quad
    \min_{\{h_k \mid 1\leq k<N,\;h_k\notin\mathcal{B}\}}
    \sum_{k=1}^{N-1} e_k(h_k,\hat{h}_k)
    + \mathcal{L}(\hat{h}_N,y)
    \label{equ:intra-chunk}
\end{equation}
Here $\hat{h}$ is the single layer prediction, and $\hat{h}_N$ is the final prediction. Only non-boundary hidden activities are optimized. Boundary activities are held fixed at the values obtained from inter-chunk inference; weights, input $x$, and target $y$ also remain fixed. The objective retains all hidden-layer errors, including errors at fixed boundaries, which constrain the preceding internal activities. The weights are then updated as in \cref{equ:weight-update}. We call this fully local instantiation PC-PC.

We also evaluate a hybrid control in which the boundary alignment stage is unchanged, but the weights inside each chunk are updated by backpropagation:
\begin{equation}
\text{Intra-chunk BP Learning}\quad W_i \leftarrow
\begin{cases}
W_i-\eta\nabla_{W_i}e_j(h_j,\hat{h}_j),
& h_j\in\mathcal{B}, \\[0.5em]
W_i-\eta\nabla_{W_i}\mathcal{L}(\hat{h}_N,y),
& j=N.
\end{cases}
\label{equ:intra-chunk-bp}
\end{equation}
Here $i\in\{1,\ldots,N\}$ indexes a layer as before, and $j$ is the index of the final layer (boundary) of the chunk containing layer $i$. The prediction $\hat{h}_j$ is obtained by a forward pass through the chunk, starting from its fixed input activity. During the weight update, the chunk input and inferred boundary target $h_j\in\mathcal{B}$ are held fixed, while gradients propagate through all layers within the chunk. The final chunk, whose endpoint is $j=N$, uses the supervised loss. We call this hybrid control PC-BP. Its role is to test whether the boundary-first schedule is useful even when intra-chunk learning is not purely predictive coding.

\section{Experiments}

Following prior PCN evaluations using fully connected architectures \cite{Song2020,Tschantz2025,Innocenti2025}, we perform experiments on MNIST and CIFAR-10 using MLP parameterizations. We train standard MLPs and $\mu$MLPs with four training dynamics: end-to-end global backpropagation, standard predictive coding, PC-PC, and the PC-BP hybrid control.

The main experiments are conducted using 20-layer MLPs with 500 hidden units per layer. For PC-PC and PC-BP, the 20 layers are divided into four chunks with five layers per chunk.
Prior work sets the number of inference steps equal to the number of hidden layers \cite{Innocenti2025}. Following this depth-based budgeting approach, we use the 20-layer depth of our main models as a reference and allocate 10 additional inference steps to ensure sufficient inference, giving a total of 30 inference steps. 
For each training dynamic, we tune the learning rate and the inference rate using 30 Optuna trials \cite{Akiba2019}. Models are trained for 5 epochs with batch size of 128. The learning rate is tuned separately for each method, and all experiments are repeated with five random seeds.

We evaluate three aspects of learning. First, we measure test accuracy to assess predictive performance. Second, we record CKA between initial and final representations to estimate the extent of feature learning. Third, we compute gradient similarity between layers to quantify feature diversity. We also vary total depth and chunk size to test whether the effect depends on a particular chunk partition.

\section{Results}

We first report predictive performance and representation change, then analyze whether the resulting updates resemble backpropagation. We finally examine feature diversity and study how total depth and chunk size affect the performance.

\subsection{Boundary-First Inference Improves Feature Learning and Performance}

\Cref{tab:acc} reports test accuracy for standard predictive coding, PC-PC, and PC-BP on MNIST and CIFAR-10. The gains on MNIST are small, consistent with the linear separability of the task. On CIFAR-10, the boundary-first schedule improves standard predictive-coding training. PC-PC increases accuracy from $41.67\%$ for standard PC to $51.44\%$ in the standard MLP setting, a gain of 9.77 percentage points. PC-BP increases accuracy from $48.77\%$ for standard PC to $54.28\%$ in the $\mu$MLP setting, a gain of 5.51 percentage points.

\begin{table}[h!]
    \caption{Test accuracy of standard backpropagation, standard predictive coding, boundary-first PC-PC, and the PC-BP hybrid control on MNIST and CIFAR-10. Values are mean $\pm$ standard deviation across five seeds. Bold indicates the highest mean in each row.}
    \label{tab:acc}
    \centering
    \resizebox{1.0\textwidth}{!}{
        
\begin{minipage}{15.8cm}
\newcommand{\meanstd}[2]{#1\,\raisebox{-0.35ex}{\scalebox{0.5}{$\pm\,#2$}}}
\centering
\sffamily
\small
\setlength{\tabcolsep}{5.5pt}
\renewcommand{\arraystretch}{1.18}

\begin{tabular}{llcccc}
\toprule
Dataset & Model & Standard BP & Standard PC & PC-PC & PC-BP \\
&&&& (Ours) & (Ours) \\
\midrule
MNIST & standard MLP & \meanstd{0.9765}{0.0041} & \meanstd{0.9686}{0.0020} & \meanstd{0.9723}{0.0006} & \textbf{\boldmath\meanstd{0.9810}{0.0008}} \\
MNIST & $\mu$MLP & \meanstd{0.9772}{0.0029} & \meanstd{0.9765}{0.0013} & \meanstd{0.9776}{0.0007} & \textbf{\boldmath\meanstd{0.9788}{0.0011}} \\
\midrule
CIFAR-10 & standard MLP & \meanstd{0.5101}{0.0049} & \meanstd{0.4167}{0.0346} & \textbf{\boldmath\meanstd{0.5144}{0.0058}} & \meanstd{0.5076}{0.0074} \\
CIFAR-10 & $\mu$MLP & \meanstd{0.5390}{0.0058} & \meanstd{0.4877}{0.0197} & \meanstd{0.5366}{0.0050} & \textbf{\boldmath\meanstd{0.5428}{0.0016}} \\
\bottomrule
\end{tabular}

\par\vspace{4pt}

\end{minipage}

    }
\end{table}

\begin{figure}[h!]
    \centering
    \includegraphics[width=0.9\textwidth]{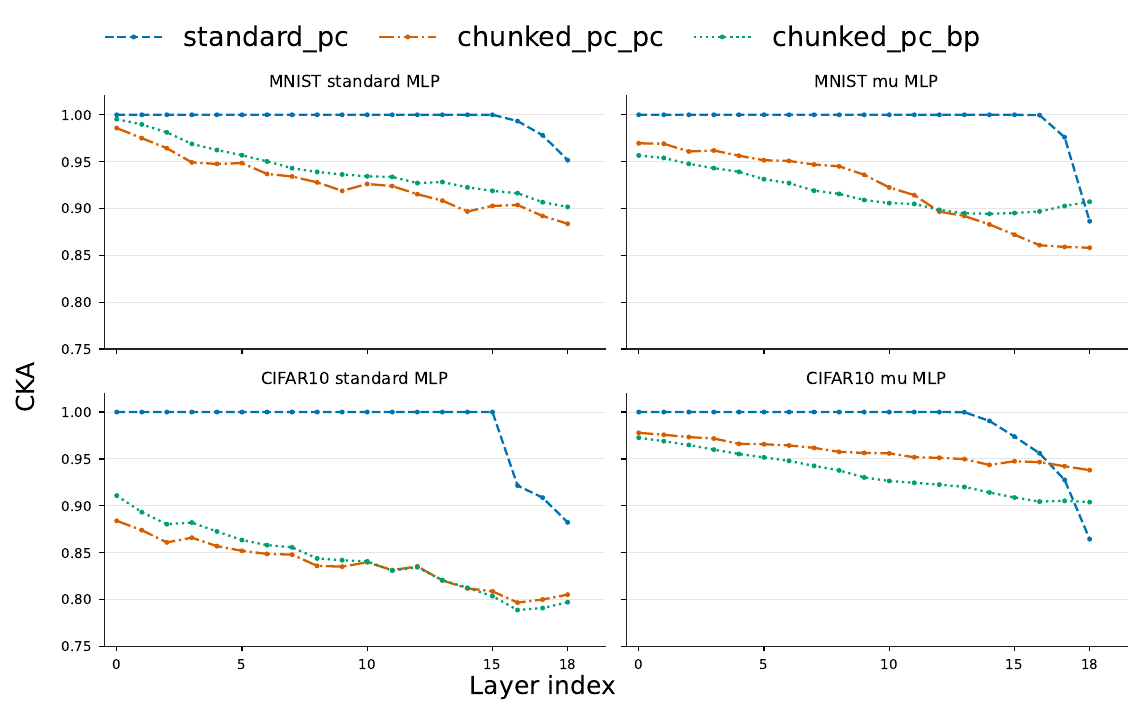}
    \caption{CKA similarity between initial and final representations for standard predictive coding, PC-PC, and PC-BP on MNIST and CIFAR-10. Lower CKA indicates greater representation change.}
    \label{fig:ini-cka}
\end{figure}

We next measure feature learning using initial-to-final CKA, as defined in \cref{equ:cka}. The results in \cref{fig:ini-cka} show that PC-PC and PC-BP reduce CKA relative to standard predictive coding training dynamics on both datasets. Lower CKA indicates greater representation change, so these results suggest that the proposed schedule mitigates the near-kernel behavior observed in standard PC.

\subsection{Boundary-First Predictive Coding Does Not Simply Imitate Backpropagation}

\begin{figure}[h!]
    \centering
    \includegraphics[width=0.9\textwidth]{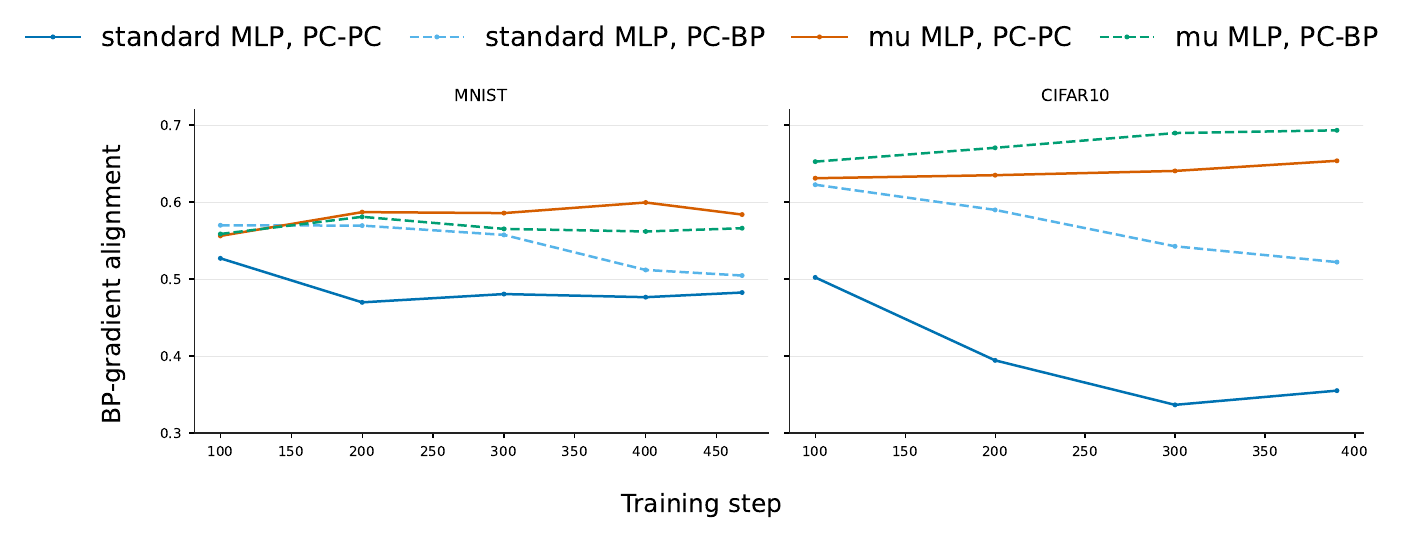}
    \caption{Gradient alignment of predictive-coding updates relative to global backpropagation.}
    \label{fig:grad-align}
\end{figure}

The performance gains could arise because the boundary-first schedule makes predictive-coding updates more similar to backpropagation. To test this hypothesis, we measure gradient alignment between predictive-coding updates and global backpropagation updates during one epoch of training. The results are shown in \cref{fig:grad-align}. The $\mu$MLP has higher alignment than the standard MLP, and PC-BP has higher alignment than PC-PC. Nevertheless, all predictive-coding variants remain below $0.7$ alignment with global backpropagation. This indicates that the boundary-first methods improve training dynamics without simply reproducing global BP updates.

\subsection{Boundary-First Inference Produces More Diverse Layerwise Updates}

\begin{figure}[h!]
    \centering
    \includegraphics[width=0.9\textwidth]{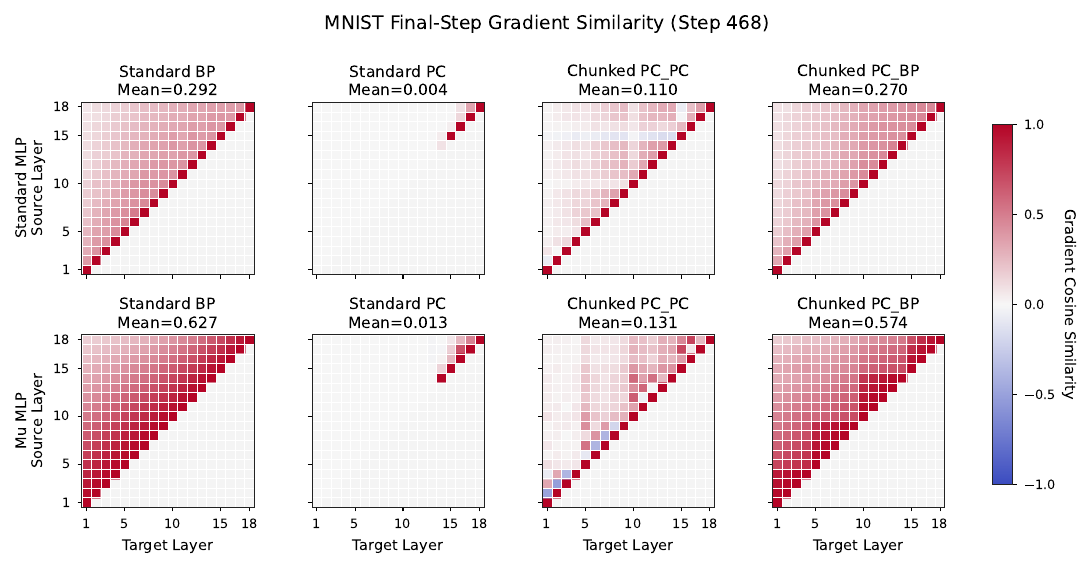}
    \caption{Gradient similarity across layers during training on MNIST. Lower similarity indicates more diverse layerwise updates. Standard predictive coding (PC) leaves several early layers with weak gradients, whereas boundary-first PCs activate more layers while maintaining a pattern of low gradient similarity.}
    \label{fig:grad-similarity-minist}
\end{figure}

We then examine whether the boundary-first schedule changes how
different layers learn. We compute the similarity between layerwise
parameter updates as
\begin{equation}
    S_{ij} =
    \frac{u_i^\top u_j}{\|u_i\|\|u_j\|+\epsilon},
\end{equation}
where $u_i$ is the flattened parameter update vector produced for layer $i$, and $\epsilon=10^{-12}$ prevents division by zero. Entries approach zero when
$\|u_i\|\|u_j\| \ll \epsilon$. \Cref{fig:grad-similarity-minist} reports gradient similarity between layers on MNIST, using the metric in \cref{equ:gradient-similarity}. All predictive coding dynamics yield lower gradient similarity than global backpropagation in the measured settings. This supports the interpretation that the boundary-first schedule does more than increase update magnitude: it also encourages different layers to move in more distinct directions. The CIFAR-10 analysis in the Appendix shows the same qualitative trend (\Cref{fig:grad-similarity-cifar}).

The more important pattern is that in standard predictive coding, early-layer gradients are close to zero for both standard MLPs and $\mu$MLPs. These layers largely transmit the input rather than learning distinct features. PC-PC and PC-BP activate these early layers and reduce the similarity between layerwise updates.

\subsection{Effects of Total Depth and Chunk Size}

\begin{figure}[h!]
    \centering
    \includegraphics[width=0.9\textwidth]{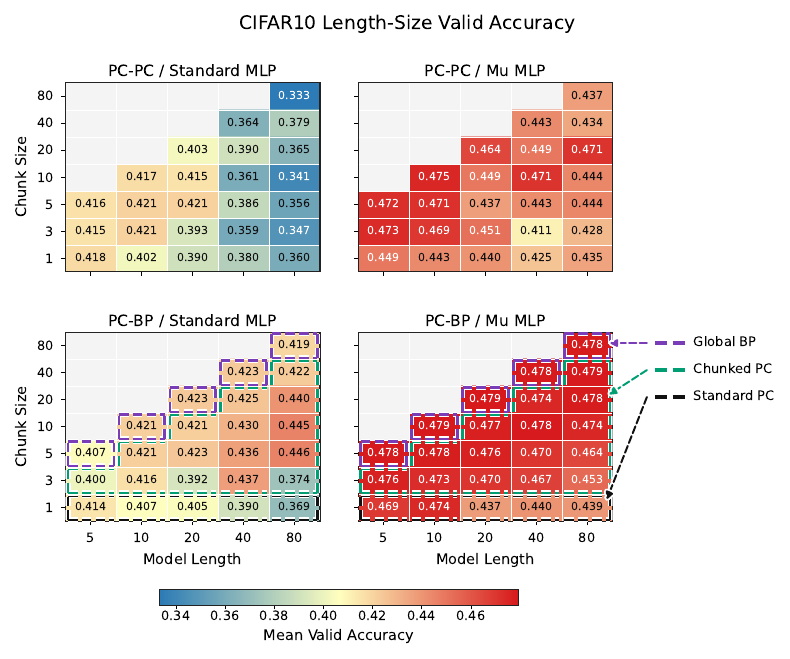}
    \caption{Mean validation accuracy for different network depths and chunk sizes on CIFAR-10. Dashed outlines and arrows indicate the endpoint correspondences for PC-BP: a single full-network chunk recovers global backpropagation.
    }
    \label{fig:layers-chunks-cifar10}
\end{figure}

Finally, we vary the model depth and chunk size on CIFAR-10 for one epoch across five random seeds.
\Cref{fig:layers-chunks-cifar10} summarizes the results for standard and $\mu$MLPs trained with PC-PC and PC-BP. For PC-BP, a chunk size of 1 recovers the standard-PC update rule, whereas a chunk size equal to the network depth recovers global backpropagation. These endpoints place standard predictive coding and global backpropagation within a single chunk-based framework as special cases. Hyperparameters are tuned independently for each configuration.

The boundary-first schedule improves upon standard predictive coding across most measured network depths and chunk sizes. The main exception is the 5-layer setting, where the benefit of boundary-first inference is limited because the network is too shallow. This pattern suggests that the schedule is most useful when the network is deep enough to learn diverse features. The results also indicate that PC-PC is more effective in smaller networks than in deeper ones, which suggests that signal propagation remains a challenge even under boundary-first optimization.

\section{Conclusion}

This study examines hidden-state optimization order through a boundary-first, chunk-based inference schedule in local learning MLPs. The schedule partitions a network into chunks, first aligns hidden states at chunk boundaries, and then refines activities within each chunk.
We instantiated this schedule in predictive coding networks, where hidden activities and local prediction errors make the optimization order operational.
Across MNIST and CIFAR-10 experiments, the boundary-first schedule improved standard predictive coding, increased early-layer update activity, reduced initial-to-final CKA, and produced more diverse layerwise gradients. These findings support boundary-first inference as a useful schedule for predictive-coding training. The evidence is empirical and specific to the PCN instantiations studied here. Extending the same optimization schedule to other local-learning frameworks, such as equilibrium propagation, architectures such as CNNs and Transformers, and other tasks remains an important direction for future work.

\section{Acknowledgment}
This work was supported by JST SPRING, Japan, under Grant Number JPMJSP2136; JSPS Grant-in-Aid for Challenging Exploratory Research under Grant Number JP25K22833; and JSPS Grant-in-Aid for Transformative Research Areas (A) under Grant Number JP22H05194. The authors would like to thank their supervisor and colleagues for their valuable feedback and support.

\bibliographystyle{unsrt}
\bibliography{PC}

\clearpage
\appendix

\section{Experimental setup}
We train 20-layer MLPs with 500 hidden units per layer on flattened MNIST and CIFAR-10 images, using dataset-specific normalization without augmentation. Models use ReLU activations without biases or dropout. The standard MLP uses residual connections and pre-layer normalization.  The $\mu$MLP uses standard Gaussian weights, residual hidden layers, and input, hidden, and output scaling factors $d_{\mathrm{in}}^{-1/2}$, $(dL)^{-1/2}$, and $d^{-1}$, respectively, where $d_{\mathrm{in}}$, $d$, and $L$ denote input dimension, hidden width, and depth. Weights are updated using Adam; global BP minimizes cross-entropy, while the predictive-coding variants use squared prediction and output errors. Hidden activities are initialized by a feedforward pass and updated synchronously by gradient descent with weights fixed. Training uses five epochs, batch size 128, and seeds $0$--$4$ controlling initialization and data ordering. Optuna evaluates 30 trials per setting over weight and activity learning rates. We also tuned the hyperparameters for each individual chunk-size combination in \cref{fig:layers-chunks-cifar10}.

\section{Feature Diversity under Boundary-First Predictive Coding on CIFAR-10}

\begin{figure}[h!]
    \centering
    \includegraphics[width=0.9\textwidth]{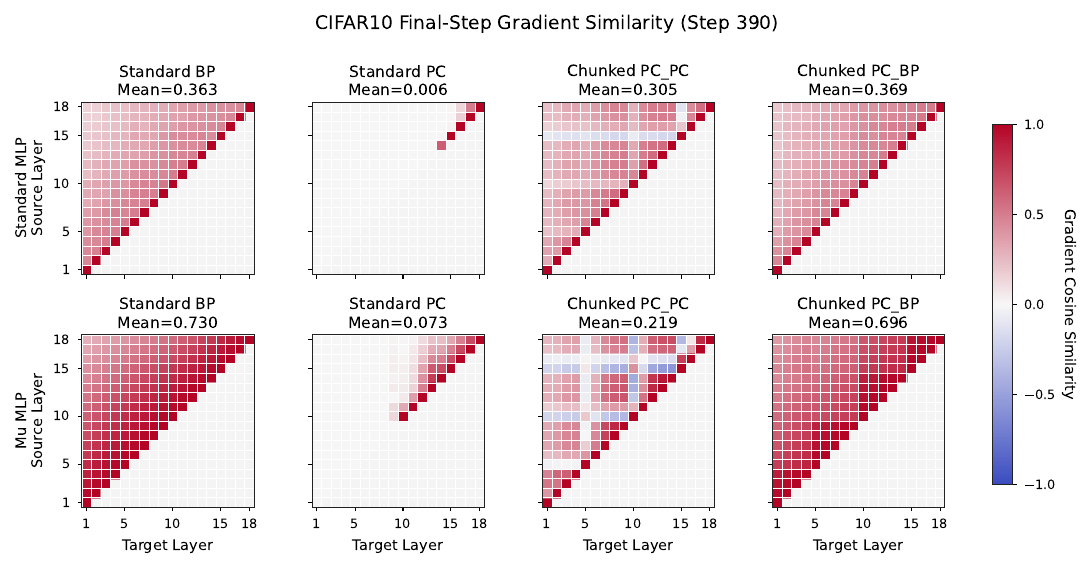}
    \caption{Gradient similarity between different layers during training on CIFAR-10. Lower similarity indicates more diverse layerwise updates.}
    \label{fig:grad-similarity-cifar}
\end{figure}

\Cref{fig:grad-similarity-cifar} shows gradient similarity between layers during training on CIFAR-10. The pattern is consistent with the MNIST results. Standard predictive coding leaves several early layers with weak gradients, while PC-PC and PC-BP activate more layers and reduce update similarity. The CIFAR-10 setting also shows that stronger input nonlinearity can increase feature learning under standard PC, but the boundary-first schedule still produces more broadly distributed layerwise updates.

\section{Details about Computing CKA}

We quantify representational change using layerwise linear CKA between the initial model and the model at the last logged training step. For each run, we construct a fixed, class-balanced test subset containing $N_{\mathrm{SAMPLES}}=100$ examples. Representations are extracted during feedforward evaluation. For each hidden layer, we form activation matrices with samples as rows and features as columns, and then center each feature across samples before applying \cref{equ:cka}. No additional feature standardization or sample normalization is applied when computing CKA. CKA is computed over the entire fixed subset for each layer, and the readout layer is excluded from the plot. \Cref{fig:ini-cka} indicates a change in representational similarity structure. We interpret this change, together with classification accuracy and update diagnostics, as supporting evidence for feature learning rather than lazy learning.

\section{Layerwise Parameter Update Norms}
\begin{figure}[h!]
    \centering
    \includegraphics[width=0.9\textwidth]{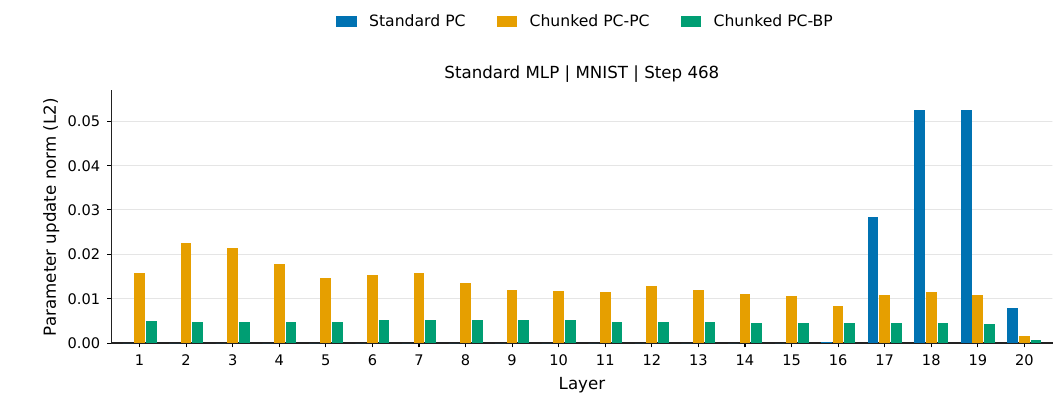}
    \includegraphics[width=0.9\textwidth]{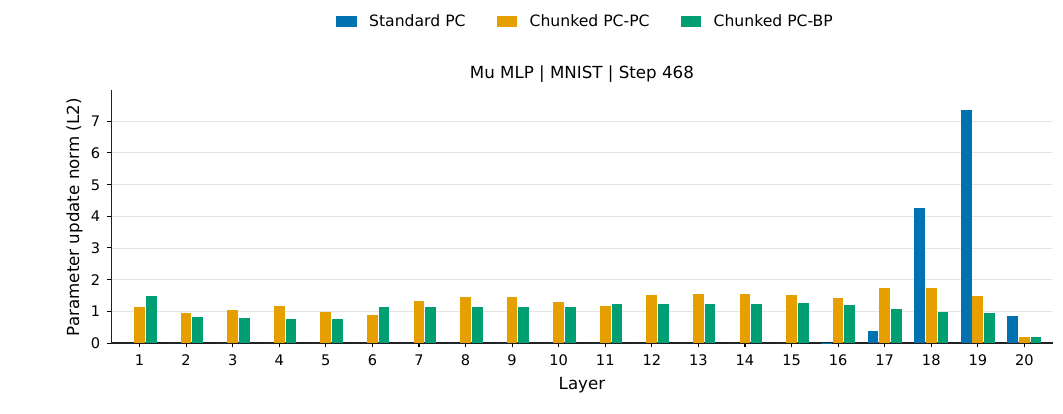}
    \caption{Layerwise L2 norms of parameter updates on MNIST for the standard MLP (top) and $\mu$MLP (bottom). These norms provide the update magnitudes needed to interpret the corresponding similarity matrices. Near-zero similarity values involving negligible updates should not be interpreted as evidence of directional diversity.}
    \label{fig:grad-norm-minist}
\end{figure}

We report the L2 norm of each layer's parameter update used in \cref{equ:gradient-similarity}. The norms are evaluated on MNIST at the same training step as the corresponding similarity matrices at \cref{fig:grad-norm-minist}.

\section{Computational Cost}

The proposed boundary-first predictive coding methods have the same asymptotic computational complexity as standard predictive coding under a fixed inference-step budget. For a dense network with approximately uniform hidden dimension $d$, the computational costs of one mini-batch training update are
\begin{equation}
\begin{aligned}
C_{\mathrm{PC}} &= O\left((T+1)BLd^2\right),\\
C_{\mathrm{PC-PC}} &= O\left((T_b+T_i+1)BLd^2\right),\\
C_{\mathrm{PC-BP}} &= O\left((T_b+1)BLd^2\right),
\end{aligned}
\end{equation}
where $T$ is the number of inference steps in standard PC, $T_b$ is the number of inter-chunk inference steps, $T_i$ is the number of intra-chunk inference steps, $B$ is the batch size, $L$ is the number of layers, and $d$ is the hidden dimension. The constant-order term represented by (+1) accounts for fixed computations such as activity initialization and weight updates. When $T=T_b+T_i$, standard PC and PC-PC therefore have the same asymptotic complexity, $O(TBLd^2)$. PC-BP has complexity $O(T_bBLd^2)$, although its actual runtime may differ because inter-chunk inference and within-chunk backpropagation involve different computational dependencies and constant factors.

\section{Pseudocode for Boundary-First Predictive Coding}
We also provide pseudocode for one minibatch update of standard PC, PC-PC, and PC-BP in \cref{fig:training-pseudocode}. 
\begin{figure}[htbp]
    \centering
    \resizebox{\textwidth}{!}{\input{standalone/training_pseudocode.tex}}
    \caption{One minibatch update for standard PC, PC-PC, and PC-BP. Standard PC infers all hidden activities together; PC-PC first infers boundaries and then interiors; PC-BP infers boundaries and then updates weights by backpropagation within each chunk.}
    \label{fig:training-pseudocode}
\end{figure}
\end{document}